\documentclass[conference]{IEEEtran}
\ifCLASSINFOpdf
\usepackage[pdftex]{graphicx}
\graphicspath{{/home/cruvadom/Desktop/temp/}}
\DeclareGraphicsExtensions{.pdf,.jpeg,.png}
\else
\fi
\usepackage{array}

\ifCLASSOPTIONcompsoc
  \usepackage[caption=false,font=normalsize,labelfont=sf,textfont=sf]{subfig}
\else
  \usepackage[caption=false,font=footnotesize]{subfig}
\usepackage{amsmath}
\usepackage{flushend}
\usepackage{url} 

\begin{document}
%
\title{Convolutional RNN: an Enhanced Model for Extracting Features from Sequential Data}

\author{\IEEEauthorblockN{Gil Keren}
\IEEEauthorblockA{Chair of Complex and Intelligent systems\\ University of Passau\\ Passau, Germany\\
cruvadom@gmail.com}
\and
\IEEEauthorblockN{Bj{\"o}rn Schuller}
\IEEEauthorblockA{Chair of Complex and Intelligent systems\\ University of Passau\\ Passau, Germany, \\ Machine Learning Group \\ Imperial College London, U.K. \\ schuller@ieee.org}}


%


\maketitle

\begin{abstract}
Traditional convolutional layers extract features from patches of data by applying a non-linearity on an affine function of the input. We propose a model that enhances this feature extraction process for the case of sequential data, by feeding patches of the data into a recurrent neural network and using the outputs or hidden states of the recurrent units to compute the extracted features. By doing so, we exploit the fact that a window containing a few frames of the sequential data is a sequence itself and this additional structure might encapsulate valuable information. In addition, we allow for more steps of computation in the feature extraction process, which is potentially beneficial as an affine function followed by a non-linearity can result in too simple features. Using our convolutional recurrent layers, we obtain an improvement in performance in two audio classification tasks, compared to traditional convolutional layers. \texttt{Tensorflow} code for the convolutional recurrent layers is publicly available in \url{https://github.com/cruvadom/Convolutional-RNN}.
\end{abstract}


%
\IEEEpeerreviewmaketitle

%

%

\section{Introduction}
Over the last years, Convolutional Neural Networks (CNN) \cite{lecun1989backpropagation} have yielded state-of-the-art results for a wide variety of tasks in the field of computer vision, such as object classification \cite{krizhevsky2012imagenet}, traffic sign recognition \cite{ciresan2012multi} and image caption generation \cite{DBLP:conf/cvpr/VinyalsTBE15}. The use of CNN was not limited to the field of computer vision, and these models were adapted successfully to a variety of audio processing and natural language processing tasks such as speech recognition \cite{sainath2015learning, DBLP:journals/corr/AmodeiABCCCCCCD15} and sentence classification \cite{kim2014convolutional}. When applying a convolutional layer on some data, this layer is extracting features from local patches of the data and often is followed by a pooling mechanism to pool values of features over neighboring patches. The extracted features can be the input of the next layer in a neural network, possibly another convolutional layer or a classifier. Models using convolutional layers for extracting features from raw data can outperform models using hand-crafted features and achieve state-of-the-art-results, such as in \cite{krizhevsky2012imagenet, DBLP:journals/corr/AmodeiABCCCCCCD15}.

Recurrent Neural networks (RNN), are models for processing sequential data which is often of a varied length such as text or sound. Long-Short Term Memory networks (LSTM) \cite{hochreiter1997long} are a special kind of recurrent neural networks that use a gating mechanism for better modeling of long-term dependencies in the data. These models have been used successfully for speech recognition \cite{DBLP:conf/icassp/GravesMH13} and machine translation \cite{cho-al-emnlp14} \cite{sutskever2014sequence}.  

When the data is a sequence of frames and of a varied length, a model combining both, convolutional layers and recurrent layers, can be constructed in the following way: the convolutional layers are used to extract features from data patches that in this case are windows comprised of a few consecutive frames in the sequence. These features extracted from windows are another time sequence, which can be the input of another convolutional layer, or finally the input of a recurrent layer that models the temporal relations in the data. One additional benefit of this method is that using a pooling mechanism after the convolutional layer can shorten the input sequence to the recurrent layer, so that the recurrent layers will need to model temporal dependencies over a smaller number of frames. 

A possible limitation of traditional convolutional layers is that they use a non-linearity applied on affine functions to extract features from the data. Specifically, an extracted feature is calculated by element-wise multiplication of a data patch by a weight matrix, summing the product over all elements of the matrix, adding a bias scalar, and applying a non-linear function. Theoretically, the function that maps a data patch to a scalar feature value can be of arbitrary complexity and it is plausible that in order to achieve a better representation of the input data, a more complicated non-linear functions can be used (i.e., with more layers of computation). Note that simply stacking more convolutional layers with kernel sizes bigger than $1\times 1$ does not solve this issue, because a consequent layer will mix the outputs of the previous layer for different locations and will not exhibit the potentially wanted behavior of extracting more complicated features from a data patch using only this data patch itself. 

In the case of sequential data, windows of a few consecutive data frames have an additional property: every window is itself a small sequence comprised of a few frames. This additional property can be potentially exploited in order to extract better features from the windows. In this work, we introduce the CRNN (Convolutional Recurrent Neural Network) model that feeds every window frame by frame into a recurrent layer and use the outputs and hidden states of the recurrent units in each frame for extracting features from the sequential windows. By doing so, our main contribution is that we can potentially get better features in comparison to the standard convolutional layers, as we use additional information about each window (the temporal information) and as the features are created from the different windows using a more complicated function in comparison to the traditional convolutional layers (more steps of computation).

We perform experiments on different audio classification tasks to compare a few variants of our proposed model to gain improvement in classification rates over models using traditional convolutional layers. The rest of the paper is organized as follows. In Section \ref{secRelated} we briefly discuss related work. In Section \ref{secModel} we describe how we extract features using recurrent layers and in Section \ref{secExperiments} we conduct experiments on audio data classification using our proposed models. In section \ref{secDiscussion} we discuss possible explanations for the observed results and conclude the work in Section \ref{secConclusions}.

%


\section{Related Work} \label{secRelated}
As described above, the feature extraction process of a standard convolutional layer can potentially be too simple, as it is simply the application of a non-linearity on an affine function of the data patch. In \cite{DBLP:journals/corr/LinCY13}, the authors address this issue by replacing the affine function for extracting features by a multilayer feedforward neural network to get state-of-the-art results on a few object classification datasets, but they only allow for a feedforward feature extraction process and not a recurrent one. 

Few other works attempt to extract features from patches of data by using the hidden states of the recurrent layer. In \cite{liang2015recurrent} the authors extract features from an image path by using a recurrent layer that takes at every time step as input this image path and the output of the last time step. This approach can result in better features as creating them using a recurrent network is a more complicated process (i.e., with more steps of computation), but still, there is no use of a sequential nature of the data patches themselves, and the process can be very computationally expensive as the same path is fed over and over to the recurrent net. 

In a recent work \cite{visin2015renet}, the authors split an image into a grid of patches, and scan every column in both directions with two recurrent networks, feeding into each recurrent network one image patch at a time. The activations of the recurrent units in both recurrent networks after processing a specific patch are concatenated and used as inputs for another two recurrent networks that now scan each row of patches in both directions. The activations of the recurrent units after processing a specific patch in the last two recurrent networks are concatenated and used as the input for the next layer or a classifier. This approach indeed uses hidden states of a recurrent network as extracted features, but there are a few major differences between this approach and our proposed approach. First, it is important to note that the usage of recurrent networks for sequential data is natural while using recurrent networks for images can be problematic because of the lack of natural sequences in the data. Another major difference is that in \cite{visin2015renet} each data path is fed in one piece to the recurrent layer and there is no usage of a sequential nature of the patches themselves. In addition, in \cite{visin2015renet} there is no option for overlapping patches to account for small translations in the data.

\section{Convolutional Long Short Term Memory} \label{secModel}
\subsection{Extracting local features} \label{sec3a}
We describe a general structure for a layer extracting (pooled) temporally local features from a data set of sequences. Using this terminology, both, traditional convolutional layers \cite{lecun1989backpropagation} and our method of Convolutional Recurrent Neural Network, (CRNN) can be described. Assume a labeled dataset where each example in the dataset is a sequence of a varied number of frames and all frames have the same fixed number of features. Denote $x$ an example from the dataset, $l$ the number of frames in the sequence $x$ and $k$ the number of features in each frame in each example in the dataset, so $x$ is of size $k \times l$. Note that for simplicity we assume each frame contains a one-dimensional feature vector of length $k$, but this can be easily generalized to the case of multidimensional data in each frame (such as video data) by flattening this data into one dimension. From the sequence $x$ we create a sequence of  windows, each window comprised of $r_1$ consecutive frames. Then, each window is of size $k \times r_1$, and we take a shift of $r_2$ frames between the starting point of two consecutive windows. Next, we apply a features extraction function $f$ on each window to get a set of $n$ features describing each window, meaning that the feature vector of size $n$ describing a window $w$ is $f(w)$. Applying $f$ on each window results in another sequence $x'$ in which every frame is represented with $n$ features. Finally, for pooling, we create windows of size $n \times p_1$ from the sequence $x'$ in the same way we created windows from the sequence $x$, with a shift of $p_2$ frames between the starting points of two consecutive windows, and we perform max-pooling across frames (for each feature separately) in each window to transform windows of size $n \times p_1$ to a vector of size $n \times 1$. After applying max-pooling to each window the resulting sequence contains (pooled) temporally local features of the sequence $x$ extracted by the function $f$ and can be fed into the next layer in the network or a classifier. 

\begin{figure*}[!t]
\centering
\subfloat[Extraction of features from a time window using a standard convolutional layer. For each feature the time window is multiplied element-wise by a different matrix. This product is summed and added to a bias term and then a non-linearity is applied. Here $\odot$ denotes an element-wise multiplication followed by summation.]{\includegraphics[width=3.3in]{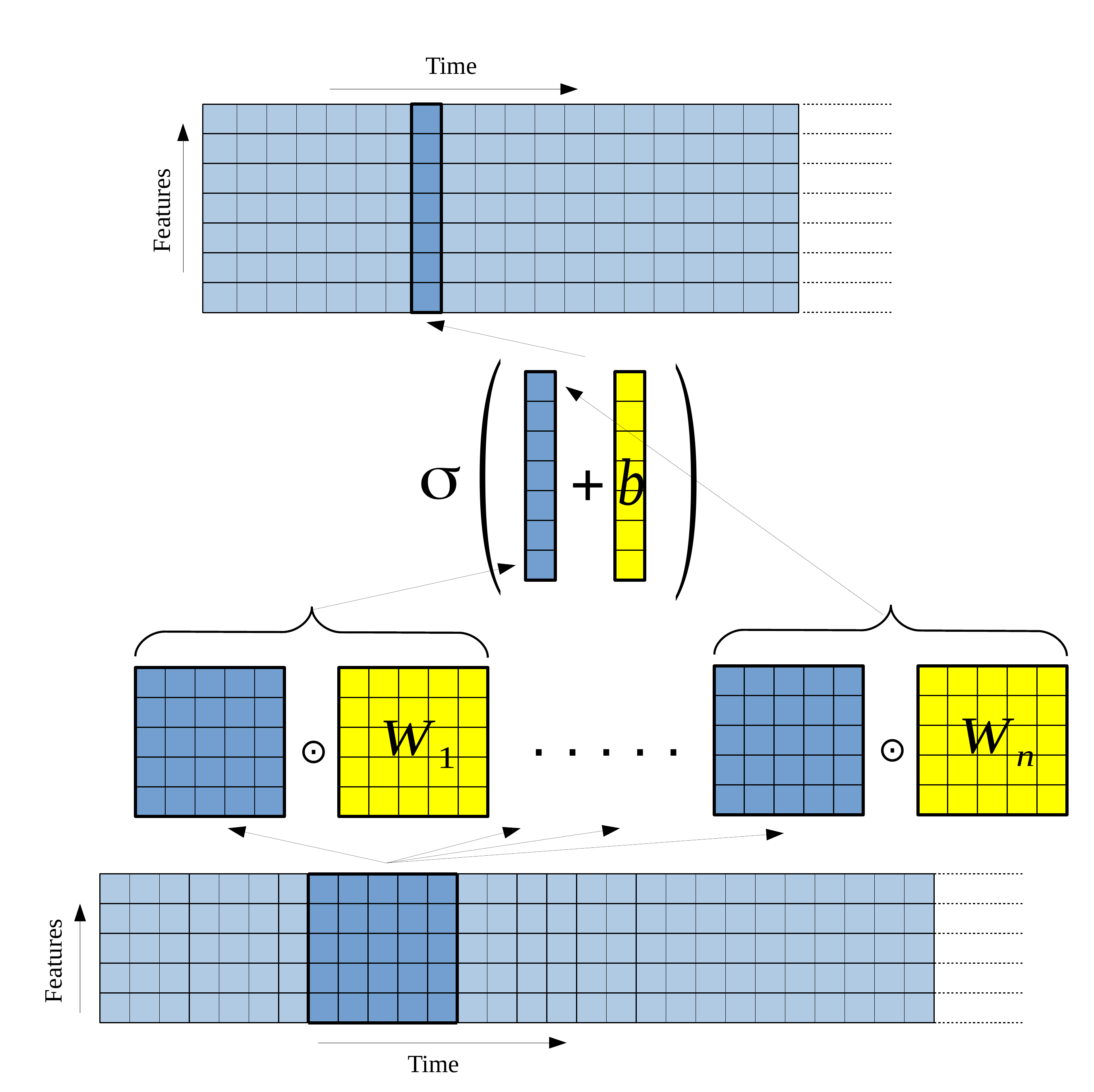}%
\label{figConv}}
\hfil
\subfloat[Extraction of features from a time window using a CRNN layer. First, the time window is fed frame by frame into a recurrent layer. Then, the hidden states of the recurrent layer along the different frames of the window are used to compute the extracted features, by applying a max/mean operator or simply by taking the vector of hidden states of the last time frame in the window. Note that it is possible to extract features from the window using the outputs of the recurrent layer in each time step instead of the hidden states.]{\includegraphics[width=3.3in]{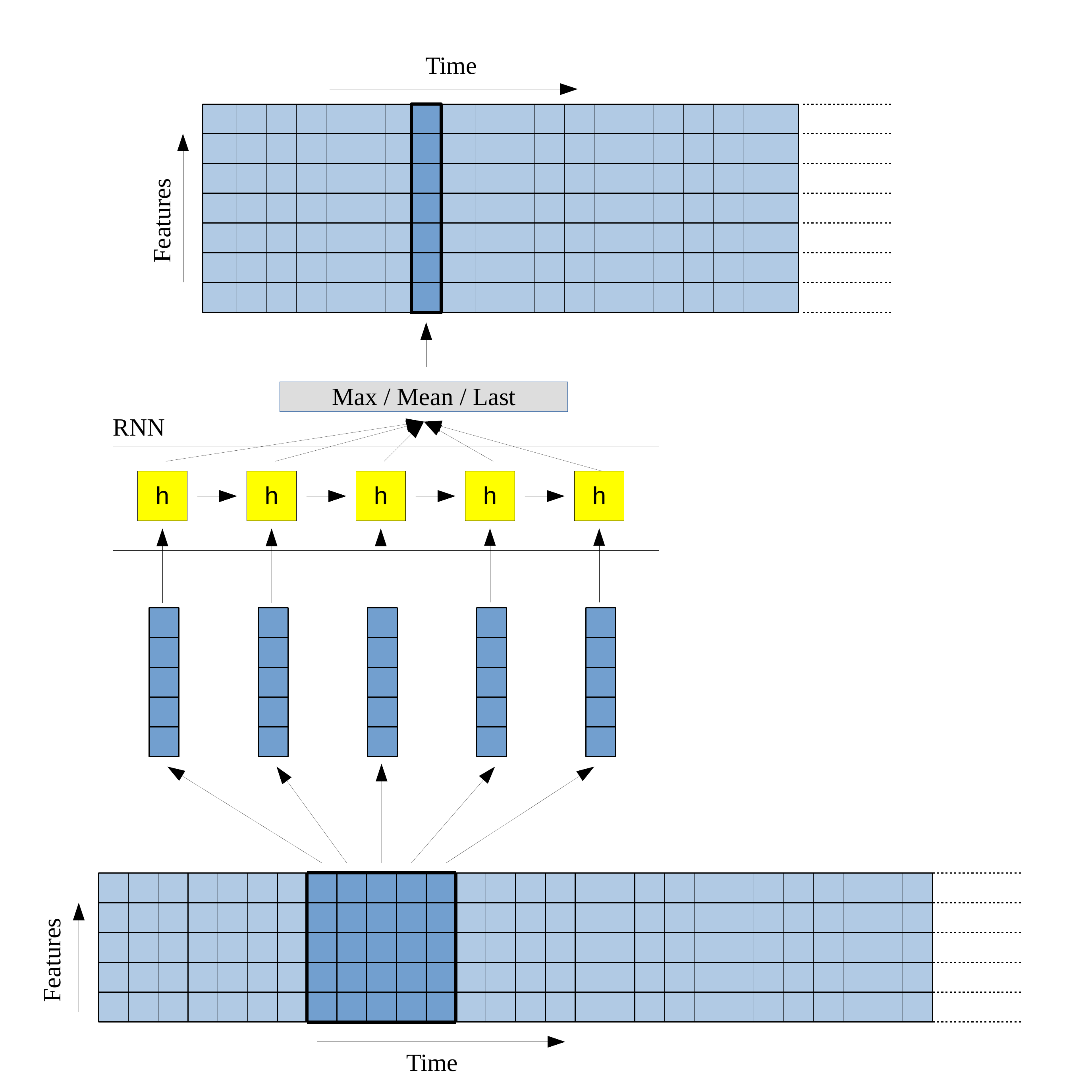}%
\label{figCRNN}}
\caption{Extraction of features from a time window by a traditional convolutional layer (a) and a CRNN layer (b).}
\label{figBig}
\end{figure*}

\subsection{Long Short Term Memory networks and Bidirectional Long Short Term Memory networks}
A simple recurrent neural network takes a sequence $(x_1, \dots ,x_t)$ and produces a sequence $(h_1, \dots ,h_t)$ of hidden states and a sequence $(y_1, \dots ,y_t)$ of outputs in the following way:
\begin{equation} \label{eqRNN}
h_t = \sigma(W_{xh}x_t + W_{hh}h_{t-1} + b_h)
\end{equation}
\begin{equation}
y_t = W_{hy}h_t + b_y,
\end{equation}
where $\sigma$ is the logistic sigmoid function, $W_{xh},W_{hh},W_{hy}$ are weight matrices and $b_h,b_y$ are biases. 

Long Short Term Memory networks \cite{hochreiter1997long} are a special kind of recurrent neural networks (RNN) that use a gating mechanism to allow better modeling of long-term dependencies in the data. The version of LSTM used in this paper \cite{gers2003learning} is implemented by replacing Equation \ref{eqRNN} with the following steps:
\begin{equation} \label{lstm1}
i_t = \sigma(W_{xi}x_t + W_{hi}h_{t-1} + W_{ci}c_{t-1} + b_i) 
\end{equation}
\begin{equation} \label{lstm2}
f_t = \sigma(W_{xf}x_t + W_{hf}h_{t-1} + W_{cf}c_{t-1} + b_f)
\end{equation}
\begin{equation} \label{lstm3}
c_t = f_t c_{t-1} + i_t tanh(W_{xc}x_t + W_{hc}h_{t-1} + b_c)
\end{equation}
\begin{equation}  \label{lstm4}
o_t = \sigma(W_{xo}x_t + W_{ho}h_{t-1} + W_{co}c_t + b_o)
\end{equation}
\begin{equation} 
h_t = o_t tanh(c_t),
\end{equation}

where $\sigma$ is the logistic sigmoid function, $i,f,o$ are the input, forget and output gates' activation vectors, and $c, h$ are cell and hidden states vectors.

Since the standard LSTM processes the input only in one direction, an enhancement of this model was proposed \cite{schuster1997bidirectional}, namely Bidirectional LSTM (BLSTM), in which the input is processed both in the standard order and reversed order, allowing to combine future and past information in every time step. A BLSTM layer comprises of two LSTM layers processing the input separately to produce $\overrightarrow{h}$, $\overrightarrow{c}$, the hidden and cell states of an LSTM processing the input in the standard order, and $\overleftarrow{h}$, $\overleftarrow{c}$, the hidden and cell states of an LSTM processing the input in reversed order. Both, $\overrightarrow{h}$ and $\overleftarrow{h}$, are then combined: 
\begin{equation} \label{eqBLSTM}
y_t = W_{\overrightarrow{h}y}\overrightarrow{h}_t + W_{\overleftarrow{h}y}\overleftarrow{h}_t + b_y,
\end{equation}
to produce the output sequence of the BLSTM layer. Note that it is possible to use the cell states instead of the hidden states of the two LSTM layers in an BLSTM layer, in the following way:
\begin{equation} \label{eqBLSTM2}
y_t = W_{\overrightarrow{c}y}\overrightarrow{c}_t + W_{\overleftarrow{c}y}\overleftarrow{c}_t + b_y,
\end{equation}
to produce the the output sequence of the BLSTM layer.

\subsection{Convolutional layers and Convolutional Recurrent layers}
Traditional convolutional layers (with max-pooling) can be described using the terms from Section \ref{sec3a} by setting the function $f$ extracting $n$ features from a data patch $w$:
\[f(w) = (\sigma(\text{sum}(W_1 \odot w) + b_1), \dots , \sigma(\text{sum}(W_n \odot w) + b_n)),\]
where $W_1,\dots ,W_n$ are weight matrices, $b_1,\dots ,b_n$ are biases, $\odot$ is an element-wise multiplication, and $sum$ is an operator that sums all elements of a matrix. This gives rise to $n$ different features computed for each window using the $n$ weight matrices and biases. The values of a specific feature across windows is traditionally called a feature map. Figure \ref{figConv} depicts the feature extraction procedure in a standard convolutional layer.

Next, we describe our proposed model, Convolutional Recurrent Neural Network (CRNN), and as special cases Convolutional Long Short Term Memory (CLSTM) and Convolutional Bidirectional Long Short Term Memory (CBLSTM), by describing the feature extraction function $f(w)$ that corresponds to these layers. A CRNN layer takes advantage of the fact that every window $w$ of size $k \times r_1$ can also be interpreted as a short time sequence of $r_1$ frames, each of size $k \times 1$. Using this fact, we can feed a window $w$ as a sequence of $r_1$ frames into an RNN. The RNN then produces a sequence of hidden states $(h_1, \dots ,h_{r_1})$ in which each element is of size $n \times 1$, where $n$ is the number of units in the RNN layer. Finally, in order to create the feature vector of length $n$ that represents a window $w$, we can either set $f(w)$ to be the mean or max over vectors of the sequence $(h_1, \dots ,h_{r_1})$ (computing mean or max for each feature separately), or simply set $f(w)$ to be the last element of the sequence. Figure \ref{figCRNN} depicts the feature creation process in a CRNN layer. As special case of CRNN we have the CLSTM and CBLSTM, in which the RNN layer is an LSTM or BLSTM, respectively. 

Note that we use the same RNN layer to compute $f(w)$ for all windows $w$, meaning that as in a convolutional layer, we use the same values of parameters of the model to extract local features for all windows. In addition, note that the RNN may produce two additional sequences, a sequence of outputs $(y_1, \dots ,y_{r_1})$ and a sequence of cell states $(c_1, \dots c_{r_1})$ (in the case of CLSTM or BLSTM) that can be used instead of the hidden states sequence to compute the features representing a window. In case we use the sequence of outputs, the hidden dimension of the recurrent layer is free of constraints.

We propose another extension for the CLTSM layer, namely the Extended CLSTM layer. In Extended CLSTM, we want to allow an LSTM layer to use the additional information about the position of each frame inside the time window. In a standard LSTM this is not possible, since all frames being fed into the LSTM are using the same weight matrices $W_{xi}, W_{xf}, W_{xc}, W_{xo}$ from equations \ref{lstm1}, \ref{lstm2}, \ref{lstm3} and \ref{lstm4}. Therefore, in an Extended CLSTM layer we use a different copy of each of the above four matrices for each frame in the window.

%
%


\section{Experiments} \label{secExperiments}
\subsection{Experiments on emotion classification: FAU-Aibo corpus} \label{secAibo}
The FAU Aibo Emotion Corpus \cite{steidl2009automatic} contains audio recordings of children interacting with Sony's pet robot Aibo. The corpus consists of spontaneous German speech where emotions are expressed. We use the same train and test subsets and labels as were used in the Interspeech 2009 Emotion Challenge \cite{schuller2009interspeech}. The corpus contains about 9.2 hours of speech, where the train set comprises of 9,959 examples from 26 children (13 male, 13 female) and the test set comprises of 8,257 examples from 25 children (8 male, 17 female). The train and test subsets are speaker-independent. A subset of the training set containing utterances from the first five speakers (according to the dataset's speaker IDs) was used as a validation set and utterances from these five speakers were not used for training the model. The dataset is labeled according to the emotion demonstrated in each utterance, and there exist two versions for labels of this dataset. In the five labels version, each utterance in the corpus is labeled with one of: \textsc{Anger}, \textsc{Emphatic}, \textsc{Neutral}, \textsc{Positive}, \textsc{Rest}. In the two labels version, each utterance in the corpus is labeled either as \textsc{Negative} or \textsc{Idle}. Since each class has a different number of instances, we add to the training set random examples from the smaller classes until the size of all classes is equal, in a way such that no example was duplicated twice before all examples from the same class were duplicated once. 

We trained the models to classify the correct emotion exhibited in each utterance. The input features used for the experiments is 26-dimensional log mel filter-banks, computed with windows of 25\,ms with a 10\,ms shift. We distinguish the term \emph{input features}, which is used to denote the input data representation (here, log mel filter-banks), from the term \emph{features} or \emph{extracted features}, which we use to describe the output of the various convolutional layers. We applied a mean and standard deviation normalization for each combination of speaker and feature independently. 

The baseline model we used contains one LSTM layer of dimension 256. For every utterance, the hidden states of the LSTM layer in each time frame were fed into a dense layer with 400 Rectified Linear Units \cite{nair2010rectified} and the outputs of the latter were fed to a softmax layer \cite{bridle1990probabilistic}. The outputs of the softmax layer in the last four time steps of each utterance were averaged, and the class with the highest averaged probability was selected as the prediction of the model. 

In addition, we evaluated models containing convolutional layers preceding the above described model, of three types: standard convolutional layers, CLSTM and Extended CLSTM. In these models, the features were first fed into two consecutive convolutional layers, each with a window size of 5 time frames, with a shift of 2 time frames between the starting points of consecutive windows. For the CLSTM and the Extended CLSTM layers, the output of the layers for each window was $c_5$, the cell states vector of the last time frame in each window. For each window, each layer extracts 100 features (therefore the dimension of the CLSTM and Extended CLSTM layers was 100). Each convolutional layer also included max-pooling over the features extracted from groups of two windows, with a shift of two windows between consecutive pooling groups (the max-pooling operator was computed for each feature separately). 
 
The parameters were learned in an end-to-end manner, meaning that all parameters of the model were optimized simultaneously, using the Adam optimization method \cite{kingma2014adam} with learning rate of $0.002$, $\beta 1=0.1$ and $\beta 2 = 0.001$ (hyperparameters of the Adam optimization method) to minimize a cross-entropy objective. Note that we did not use Back Propagation Thorough Time (BPTT) \cite{werbos1990backpropagation} and we used only the outputs of the last four time steps to compute the model predictions. This approach can potentially result in a difficulty in optimizing, as we back propagate through a very deep network. This potential problem is alleviated by the fact that using two convolutional layers makes the output sequence shorter and therefore helps overcoming this obstacle, and we did not observe any exceptional difficulties in learning the model in an end-to-end manner. 

The selected model used on the test set is the one with best misclassification rate on the validation set, and training was stopped after 12 epochs of training without improvement of the misclassification rate of the validation set. The experiments were performed using the deep learning framework Blocks \cite{van2015blocks} based on Theano \cite{Bastien-Theano-2012, bergstra+al:2010-scipy}. Results on the test set using the baseline model and using the different types of convolutional layers are reported in Table \ref{tabAibo}. Since the number of utterances from each class may vary, we report the Unweighted Average (UA) Recall, which is the official benchmark measure on this task. The UA Recall is defined to be
\[\frac{1}{n} \sum \limits_{i=1}^n r_i,\]
where $n$ is the number of classes, and the recall $r_i$ is the number of correctly classified examples from class $i$ (true positives) divided by the total number of examples from class $i$. As seen in the results, our models outperform both the baseline and traditional convolutional layers for this task, as well as the challenge's baseline \cite{schuller2009interspeech}.

\begin{table}[!t]
\renewcommand{\arraystretch}{1.3}
\caption{Results on emotion classification task (best results in bold)}
\label{tabAibo}
\centering
\begin{tabular}{|c||c|c|}
\hline
& \multicolumn{2}{c|}{Test set UA Recall [\%]} \\
\cline{2-3}
 & 2 labels &  5 labels\\
\hline
baseline & $69.01 \pm 0.93$ & $36.89 \pm 1.8$ \\
\hline
standard convolutional layer & $69.13 \pm 0.87$ & $38.22 \pm 1.19$ \\
\hline
CLSTM & \textbf{70.59} $\pm$ \textbf{0.58} & $39.41 \pm 0.57$ \\
\hline
Extended CLSTM & $69.85 \pm 0.38$ & \textbf{39.72} $\pm$ \textbf{0.2} \\
\hline
\end{tabular}
\end{table}

\subsection{Experiments on age and gender classification: aGender corpus} \label{secAGender}
The aGender corpus \cite{burkhardt2010database} contains audio recordings of predefined utterances and free speech produced by humans of different age and gender. Each utterance is labeled as one of four age groups: \textsc{Child}, \textsc{Youth}, \textsc{Adult}, \textsc{Senior}, and as one of three gender classes: \textsc{Female}, \textsc{Male} and \textsc{Child}. For the train and validation sets, we split the train set from the Interspeech 2010 Paralinguistic Challenge \cite{schuller2010interspeech} which contains 32,527 utterances in 23.43 hours of speech by 471 speakers. We split this set by selecting random 20 speakers (speaker IDs can be obtained from www.openaudio.eu) from each one of the seven groups (child, youth male, youth female, adult male, adult female, senior male, senior female) and using all utterances from these 140 speakers as a validation set. Utterances from the rest of the speakers were used as the train set. The test set we used is the set used as the development set in the Interspeech 2010 Paralinguistic Challenge and contains 20,549 utterances in 14.73 hours of speech by 299 speakers. The train, validation and test sets are speaker independent (i.e., each speaker appears in exactly one of the sets). We use this data for two tasks: age classification and gender classification. For each task, we balance the number of instances between the different classes in the same way we did with the FAU-Aibo dataset.

In these experiments with age and and gender classification, we evaluated our models using two different sets of input features. The first set of input features used is 26-dimensional log mel filter-banks, computed with windows of 25\,ms with a 10\,ms shift. The second set of input features is the extended Geneva Minimalistic Acoustic Parameter Set (eGeMAPS) \cite{eyben2015geneva}, a input features set suited for voice research and specifically paralinguistic research, containing 25 LLDs such as pitch, jitter, center frequency of formants, shimmer, loudness, alpha ratio, spectral slope and more. The eGeMAPS input features were extracted using the openSMILE toolkit \cite{eyben2013recent}, again with 25\,ms windows with a 10\,ms shift. For each feature set we applied mean and standard deviation normalization for each combination of speaker and feature independently.

The models we evaluated contain a few differences compared to the models used for emotion recognition. Our baseline model contains one BLSTM layer that allows to use past and future information in each time step. Inside the BLSTM layer, there are two LSTM layers (forward and backward) of dimension 256 each. As in Equation \ref{eqBLSTM}, in every time step the hidden states of the two LSTM layers were fed into a dense layer of dimension 400, followed by the Rectified Linear non-linearity, that in turn was fed into a softmax layer. The outputs of the softmax layer were averaged across all time steps, and the class with the highest probability was selected as the prediction of the model. 

We also evaluated models with one convolutional layer preceding the above described model, of three types: standard convolutional layer, CLSTM and CBLSTM, with window sizes, pooling sizes, shifts and number of extracted features identical to the layers in the emotion recognition experiments in Section \ref{secAibo}. The output of the CLSTM layer for each window was $max(c_1,\dots ,c_5)$ where the $max$ operator is computed for each feature separately. Inside the CBLSTM layer, each LSTM layer was of dimension 100 and the cell states of the two LSTM layers were combined to produce an output $y_t$ of size 100 in each time step $t$ as in Equation \ref{eqBLSTM2}. The output of the CBLSTM layer for each window was then $max(y_1,\dots ,y_5)$ where the $max$ operator is computed for each feature separately. The rest of the training setting is identical to the emotion recognition experiments. Again, we did not use BPTT \cite{werbos1990backpropagation}. The fact that in these experiments we used all time steps to compute the prediction of the model makes it easier to learn the model in an end-to-end manner, as it creates a shorter path between the input and the output of the model. 

Unweighted Average (UA) Recall on our test set is reported in tables \ref{tabGender} and \ref{tabAge}, which is the official benchmark measure on this task. As seen in the results, the model with the CBLSTM layer yields an improvement over the baseline, traditional convolutional and CLSTM layers for this task, using both sets of input features. In addition, in most cases the model with the CLSTM layer outperforms our baseline and the standard convolutional layer, as well as the challenge's baseline \cite{schuller2010interspeech} (for the the challenge's baseline, using the eGeMAPS input features).
 
\begin{table}[!t]
\renewcommand{\arraystretch}{1.3}
\caption{Results on gender classification (best results in bold)}
\label{tabGender}
\centering
\begin{tabular}{|c||c|c|}
\hline
& \multicolumn{2}{c|}{Test set UA Recall [\%]} \\
\cline{2-3}
 & log mel filter-banks &  eGeMAPS \\
\hline
baseline & $70.94 \pm 1.0$ & $76.96 \pm 0.4$ \\
\hline
standard convolutional layer & $74.49 \pm 0.21$ & $77.54 \pm 0.3$ \\
\hline
CLSTM & $75.68 \pm 0.7$ & $78.17 \pm 0.33$ \\
\hline
CBLSTM & \textbf{76.03} $\pm$ \textbf{1.28} & \textbf{78.23} $\pm$ \textbf{0.49} \\
\hline
\end{tabular}
\end{table}

\begin{table}[!t]
\renewcommand{\arraystretch}{1.3}
\caption{Results on age classification (best results in bold)}
\label{tabAge}
\centering
\begin{tabular}{|c||c|c|}
\hline
& \multicolumn{2}{c|}{Test set UA Recall [\%]} \\
\cline{2-3}
 & log mel filter-banks &  eGeMAPS \\
\hline
baseline & $43.14 \pm 0.94$ & $46.51 \pm 0.62$ \\
\hline
standard convolutional layer & $45.93 \pm 0.52$ & $47.09 \pm 0.76$ \\
\hline
CLSTM & $45.51 \pm 0.27$ & $47.15 \pm 0.57$ \\
\hline
CBLSTM & \textbf{46.39} $\pm$ \textbf{0.14} & \textbf{47.29} $\pm$ \textbf{0.58} \\
\hline
\end{tabular}
\end{table}

\subsection{Comparing Different CLSTM architectures}
When implementing a CLSTM model, one has two choices to make regarding the specific architecture. First, for feature extraction, one can use the sequence $(h_1,\dots , h_n)$ of hidden states, the sequence $(c_1,\dots ,c_n)$ of cell states or the sequence $(y_1,\dots ,y_n)$ of outputs. The second choice to be made is whether to apply the max or mean operators on the selected sequence (applied to each feature independently) or simply to use the last element of the selected sequence as the extracted features. Using the $max$ operator over a group of objects as the output of a layer seems to yield good results in Deep Learning in different situations such as convolutional layers with max-pooling \cite{boureau2008sparse} \cite{jarrett2009best} or Maxout Networks \cite{goodfellow2013maxout}. The $max$ operator has some desired theoretical properties as well. For example, in \cite{goodfellow2013maxout} the authors show that applying the $max $ operator over a sufficiently high number of linear functions can approximate an arbitrary convex function. Usage of the $mean$ operator has been made in Deep Learning as well, when applying mean pooling with convolutional layers \cite{lecun1998gradient}. 

We empirically evaluated six different implementations of the CLSTM model, by using the two sequences $(h_1,\dots , h_n)$ and $(c_1,\dots ,c_n)$ of hidden and cell states, each with the three operations: $max$, $mean$ and using the last. We compared the different implementations of the CLSTM model both on the emotion classification task and the age and gender classification task. For both tasks, as introduced in Section \ref{secAibo} and Section \ref{secAGender}, we used the 26-dimensional log mel filter-banks as the input features. For each task we used the same network architecture and training procedure as was used in the above experiments, with the only difference being the implementation of the CLSTM layers (as in the experiments above, the emotion classification model contains two CLSTM layers, and the age and gender classification model contains one CLSTM layer). Table \ref{tabCLSTM} contains the result for these experiments. We observed that in all four tasks, the average results of implementations using the cell states is better than the average results of implementations using the hidden states, but still in one task of the four the implementation yielding the best result used the hidden states. The results regarding which one of the six different implementations is preferable are ambiguous, and we cannot point out one implementation which is better than the others. In addition, comparing the usage of $max$, $mean$ and the last vector of a sequence also gave ambiguous results, with a slight preference towards using the $max$ operator.

\begin{table*}[!t]
\renewcommand{\arraystretch}{1.3}
\caption{Different Implementations of CLSTM (best results in bold). Details in the text.}
\label{tabCLSTM}
\centering
\begin{tabular}{|c||c|c|c|c|c|}
\hline
& \multicolumn{5}{c|}{Test set UA Recall [\%]} \\
\cline{2-6}
 & Emotion 2 classes & Emotion 5 classes & Age & Gender & Average \\
\hline
hidden states, mean & $70.04 \pm 0.37$ & $39.47 \pm 0.74$ & $45.62 \pm 1.08$ & $74.54 \pm 0.55$ & $57.42$\\
\hline
hidden states, max & $70.2 \pm 0.46$ & $39.46 \pm 1.06$ & $45.89 \pm 1.13$ & $75.02 \pm 1.23$ & $57.64$\\
\hline
hidden states, last & $70.06 \pm 0.3$ & \textbf{40.47} $\pm$ \textbf{1.02} & $45.53 \pm 0.53$ & $74.73 \pm 1.55$ & $57.7$\\
\hline
cell states, mean & $70.15 \pm 0.43$ & $40.27 \pm 1.29$ & \textbf{46.2} $\pm$ \textbf{0.33} & $75.26 \pm 0.14$ & \textbf{57.97}\\
\hline
cell states, max & $70.51 \pm 0.18$ & $39.99 \pm 0.76$ & $45.51 \pm 0.27$ & \textbf{75.68} $\pm$ \textbf{0.7} & $57.92$\\
\hline
cell states, last & \textbf{70.59} $\pm$ \textbf{0.58} &  $39.41 \pm 0.57$ & $45.67 \pm 0.72$ & $74.79 \pm 1.11$ & $57.62$\\
\hline
\hline
mean& $70.10$ & $39.87$  & \textbf{45.91}  & $74.90$ & $57.70$\\
\hline
max & \textbf{70.35} & $39.73$ & $45.70$  & \textbf{75.35} & \textbf{57.78}\\
\hline
last& $70.33$ & \textbf{39.94}  & $45.61$  & $74.76$ & $57.66$\\
\hline
\hline
hidden states & $70.10$ & $39.80$  & $45.69$  & $74.76$ & $57.59$\\
\hline
cell states & \textbf{70.42} & \textbf{39.89} & \textbf{45.79}  & \textbf{75.24}  & \textbf{57.84}\\
\hline
\end{tabular}
\end{table*}

\section{Discussion} \label{secDiscussion}
We observed that the CBLSTM outperformed the CLSTM layer in classifying age and gender of speakers. This result can have several explanations. First, when feeding a window to a CBLSTM layer, both past and future information can be used in each time step, a property that does not exist in the CLSTM layer, which processes the window in one direction only. The future information might be valuable when analyzing audio data and in particular when classifying the age and gender of a speaker, and this might explain why the CBLSTM layers performed better compared to the CLSTM layers. Another phenomenon that might help explaining the difference between CBLSTM and CLSTM is the following: when we used a CLSTM layer, we computed features for a window by applying a $max$ operator on the sequence of cell states. The cell states then have to play a double role: on the one hand, they have to contain valuable data to feed to the next time step and on the other hand, they have to contain valuable data to feed the next layer or the classifier. These two requirements can potentially contradict each other and this may hamper the performance of a CLSTM model. When we used the CBLSTM layers, we combined the cell activations of the forward and backward LSTM layers using weight matrices to compute the output $y_t$ in each time step as in Equation \ref{eqBLSTM2}, and we used $y_t$ to compute the extracted features for the window. By doing that, we achieve two things: we allow for an additional transformation on the cell states before using them for computing the extracted features, and we get a separation of the two potentially contradicting roles of the cell states described above.  

In recent years, some works perform audio related tasks such as speech recognition using the raw waveform \cite{sainath2015learning,golik2015convolutional,hoshenspeech}. For these models, the sequences can be relatively long and convolutional layers might be applied on windows containing many time steps. In that case, the CRNN model can be especially useful, because of two reasons: first, longer windows encapsulate more temporal structure that might contain valuable information. Second, while in traditional convolutional layers the number of parameters grows with the size of window, in the CRNN model the number of parameters does not depend on the length of the window (except in the extended CLSTM model) but only on the number of features per time step and the inner dimension of the CRNN layer.

Another observation from the results is that the hand-crafted input features set eGeMAPS performed better than the low complexity log mel filter-banks. This is in contrary to results from the field of computer vision and speech recognition, where in the recent years features extracted from raw data by convolutional layers outperform hand-crafted input features and achieve state-of-the-art results in various tasks. We conjecture that with bigger datasets and further development of models, simpler input features will outperform hand-crafted input features for the tasks we experimented with.

\section{Conclusion} \label{secConclusions}
We have proposed the CRNN model, to improve the process of feature extraction from windows of sequential data that is being performed by traditional convolutional networks, by exploiting the additional temporal structure that each window of data encapsulates. The CRNN model extracts the features by feeding a window frame-by-frame to a recurrent layer, and the hidden states or outputs (or cell states in case of an LSTM) of the recurrent layer are then used to compute the extracted features for the window. By doing so, we also allow for more layers of computation in the feature generation process, which is potentially beneficial. We experimented with three types of the CRNN model, namely CLSTM, Extended CLSTM and CBLSTM. We found that these models yield improvements in classification results compared to the traditional convolutional layers for the same number of extracted features. 

The improvements in classification results were observed when the input data comprised of log mel filter-banks, which are features of low complexity, and even when the input data comprised of higher level properties of the audio signal (the eGeMAPS features), the model could still extract valuable information from the data and an improvement in classification result was obtained.

In future work, futures extracted by the CRNN model can be further evaluated and compared to features extracted by standard convolutional layers. In addition, the CRNN model can be used in larger models and for various applications, such as speech recognition and video classification.

\section*{Acknowledgment}
This work has been supported by the European Community's Seventh Framework Programme through the ERC Starting Grant No.\thinspace 338164 (iHEARu). We further thank the NVIDIA Corporation for their support of this research by Tesla K40-type GPU donation.



\bibliographystyle{IEEEtran}
\bibliography{reference}
\end{document}
